\documentclass[11pt,a4paper]{article}
\usepackage[hyperref]{acl2019}
\usepackage{times}
\usepackage{latexsym}

\usepackage{url}
\usepackage{float}
\usepackage{tabularx, booktabs}
\usepackage{xspace}
\usepackage{todonotes}
\usepackage{rotating}
\usepackage{soul}
\usepackage{enumitem}
\aclfinalcopy 
\def\aclpaperid{62} 

\newcommand{\nometa}{\texttt{gen-key}\xspace}

\newcommand{\transformer}{\texttt{Transformer}\xspace}

\newcommand{\bleu}{\texttt{BLEU}\xspace}

\newcommand{\rouge}{\texttt{ROUGE-L}\xspace}

\newcommand{\ed}{\texttt{ED}\xspace}
\newcommand{\ngram}{\mbox{$n$-gram}\xspace}

\newcommand{\traingen}{\texttt{train-gen}\xspace}
\newcommand{\devgen}{\texttt{val-gen}\xspace}
\newcommand{\testgen}{\texttt{test-gen-pheno}\xspace}
\newcommand{\testgentemp}{\texttt{test-gen-temp}\xspace}
\newcommand{\original}{\texttt{all}\xspace}
\newcommand{\reduced}{\texttt{reduced}\xspace}

\newcommand{\devclass}{\texttt{dev-pheno}\xspace}
\newcommand{\testclass}{\texttt{test-pheno}\xspace}

\newcommand{\overlap}{\texttt{OVERLAP}\xspace}

\newcommand{\devtemp}{\texttt{dev-temp}\xspace}
\newcommand{\testtemp}{\texttt{test-temp}\xspace}

\newcommand{\real}{\texttt{real}\xspace}
\newcommand{\gen}{\texttt{gen}\xspace}

\title{Is artificial data useful for biomedical Natural Language Processing algorithms?}

\author{Zixu Wang$^{1}$, Julia Ive$^{2}$, Sumithra Velupillai$^{3}$ \and Lucia Specia$^{1}$\\
$^{1}$Department of Computing, Imperial College London, UK \\
$^{2}$DCS, University of Sheffield, UK\\
$^{3}$IoPPN, King's College London, UK and KTH, Sweden\\
{\tt zixu.wang18@imperial.ac.uk} \\
{\tt j.ive@sheffield.ac.uk}\\
{\tt sumithra.velupillai@kcl.ac.uk}\\
{\tt l.specia@imperial.ac.uk}}

\date{}
\begin{document}
\maketitle
\begin{abstract} 
A major obstacle to the development of Natural Language Processing (NLP) methods in the biomedical domain is data accessibility. This problem can be addressed by generating medical data artificially. 
Most previous studies have focused on the generation of short clinical text, and evaluation of the data utility has been limited.
We propose a generic methodology to guide the generation of clinical text with key phrases. We use the artificial data as additional training data in two key biomedical NLP tasks: text classification and temporal relation extraction. We show that artificially generated training data used in conjunction with real training data can lead to performance boosts for data-greedy neural network algorithms. We also demonstrate the usefulness of the generated data for NLP setups where it fully replaces real training data.
\end{abstract}

\section{Introduction}\label{sec:intro} 

Data availability is a major obstacle in the development of more powerful Natural Language Processing (NLP) methods in the biomedical domain. In particular, current state-of-the-art (SOTA) neural techniques used for NLP rely on substantial amounts of training data. 

In the NLP community, this low-resource problem is typically addressed by generating complementary data artificially \cite{poncelas:eamt:2018,edunov2018understanding}. This approach is also gaining attention in biomedical NLP.
Most of these studies present work on the generation of short text (typically under 20 tokens), given structural information to guide this generation (e.g., chief complaints using basic patient and diagnosis information~\cite{lee2018nlg}). 
Evaluation scenarios for the utility of the artificial text usually involve a single downstream NLP task (typically, text classification). 

SOTA approaches tackle other language generation tasks by applying neural models: variations of the encoder-decoder architecture (\ed) model \cite{sutskever2014,bahdanau:ICLR:2015}, a.k.a sequence to sequence (seq2seq), e.g., the \transformer model \cite{vasvani:nips:2018}. In this work, we follow these approaches and guide the generation process with key phrases in the \transformer model.

Our main contribution is thus twofold: (1)~a single methodology to generate medical text for a series of downstream NLP tasks; (2)~an assessment of the utility of the generated data as complementary training data in two important biomedical NLP tasks: text classification (phenotype classification) and temporal relation evaluation. Additionally, we thoroughly study the usefulness of the generated data in a set of scenarios where it fully replaces real training data.

\section{Related Work}\label{sec:related}

\par{\textbf{Natural Language Generation.}}
Natural language generation is an NLP area with a range of applications such as dialogue generation, question-answering, machine translation (MT), summarisation, simplification, storytelling, etc. 

SOTA approaches attempt to solve these tasks by using neural models. One of the most widely used models is the encoder-decoder architecture (\ed) \citep{sutskever2014,bahdanau:ICLR:2015}. In this architecture, the decoder is a conditional language model. It generates a new word at a timestep taking into account the previously generated words, as well as the information provided by the encoder (a sequence of hidden states, roughly speaking, a set of automatically learned features). 
 
For different tasks, the input to the encoder may be different: questions for question-answering, source text for MT, story prompts for story generation, etc.

\par{\textbf{Long text generation.}} One of the main challenges of the \ed architecture remains the generation of long coherent text. In this work, we consider paragraphs as long text. Other NLP tasks may target documents, or even group of documents (e.g., multi-document summarisation systems).

Existing vanilla \ed models mainly focus on local lexical decisions which limits their ability to model the global integrity of the text. This issue can be tackled by varying the generation conditions: e.g., guiding the generation with prompts \cite{fan:acl:2018}, with named entities \cite{clark:naacl:2018} or template-based generation \cite{wiseman2018learning}. All these conditions serve as binding elements to relate generated sentences and ensure the cohesion of the resulting text.

In this work, we follow the approach of \citet{peng:naacl:2018} and guide the generation of Electronic Health Record (EHR) notes with the help of key phrases (phrases composed of frequent content words often co-occurring with other content words). These key phrases are sense-bearing elements extracted at the paragraph level. Using them as guidance ensures semantic integrity and relevance of the generated text. We experiment with the SOTA \ed \transformer model. The model is based on multi-head attention mechanisms. Such mechanisms decide which parts of input and previously generated output are relevant for the next generation decision. Heads are designed to attend to information from different representation subspaces. Recent studies show that their roles are potentially linguistically intepretable: e.g., attending to syntactic dependencies or rare words \citep{voita-etal:2019a:ACL}.

\par{\textbf{Usage of artificial data in NLP}} 
In MT, artificial data has been successfully used in addition to real data for training \ed models. There have also been attempts to build MT models in low-resource conditions only with artificial data \citep{poncelas:eamt:2018}. In this work, we investigate the usefulness of the generated data both in the complementary setting and in the full replacement setting.

\par{\textbf{Medical text generation.}} The generation of medical data destined to help clinicians has been addressed
e.g. through generation of imaging reports by \citet{jing:acl:2018,liu:arxiv:2018}.

However, to our knowledge, there have been very few attempts to create artificial medical data to help NLP. One attempt to create such data can be found in~\citep{suominen:medinf:2015}, where nursing handover data is generated in a very costly way with the help of a clinical professional who wrote imaginary text. 

The attempt closest to ours is the one of \citet{lee2018nlg}. They generate short-length (under 20 tokens) chief complaints using diagnosis and patient- and admission-related information as conditions in the conditional LM. The authors investigate the clinical validity of the generated text by using it as test data for NLP models built with real data. But they do not look into the utility of the generated data for building NLP models.

\section{Methodology}\label{sec:method} 

\begin{figure}[h]
  \centering
  \includegraphics[width=0.7\hsize]{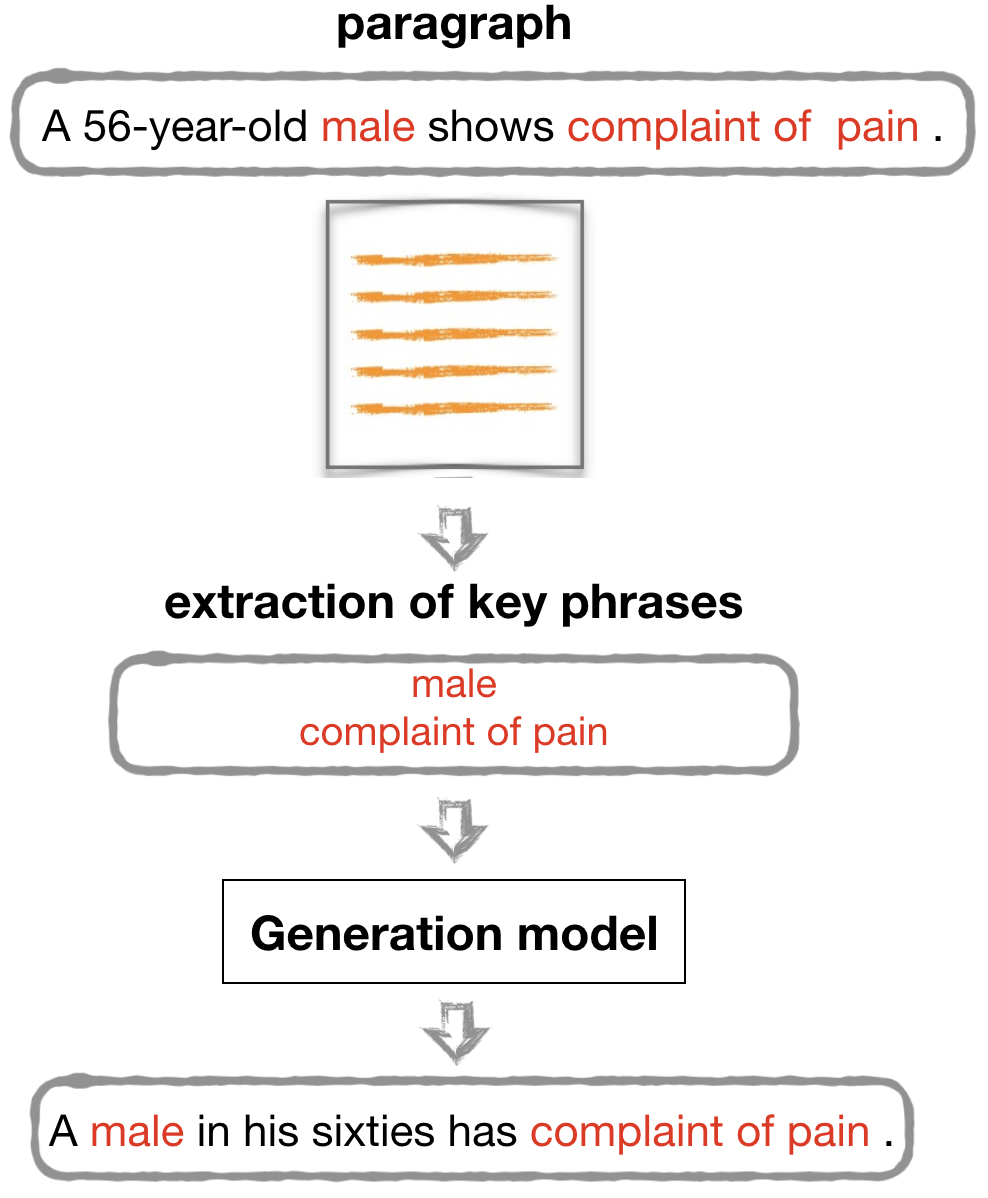}
  \caption{Our generation methodology to guide the generation with key phrases.}
  \label{fig:methodology}
\end{figure}

As mentioned before, in our attempt to find an optimal way to generate synthetic EHRs we experiment with the \transformer architecture. We extract key phrases at the paragraph level, match them at the sentence level and further use them as inputs into our generation model (see Figure~\ref{fig:methodology}). Thus, each paragraph is generated sentence by sentence but taking the information ensuring its integrity into account.

The intrinsic evaluation of the generated data is performed with a set of metrics standard for text generation tasks: \rouge~\citep{Lin:2004} and \bleu~\cite{Papieni02bleu}. \rouge measures the \ngram recall, \bleu -- the \ngram precision. We also assess the length of the generated text.

\begin{figure}[h]
  \centering
  \includegraphics[width=0.95\hsize]{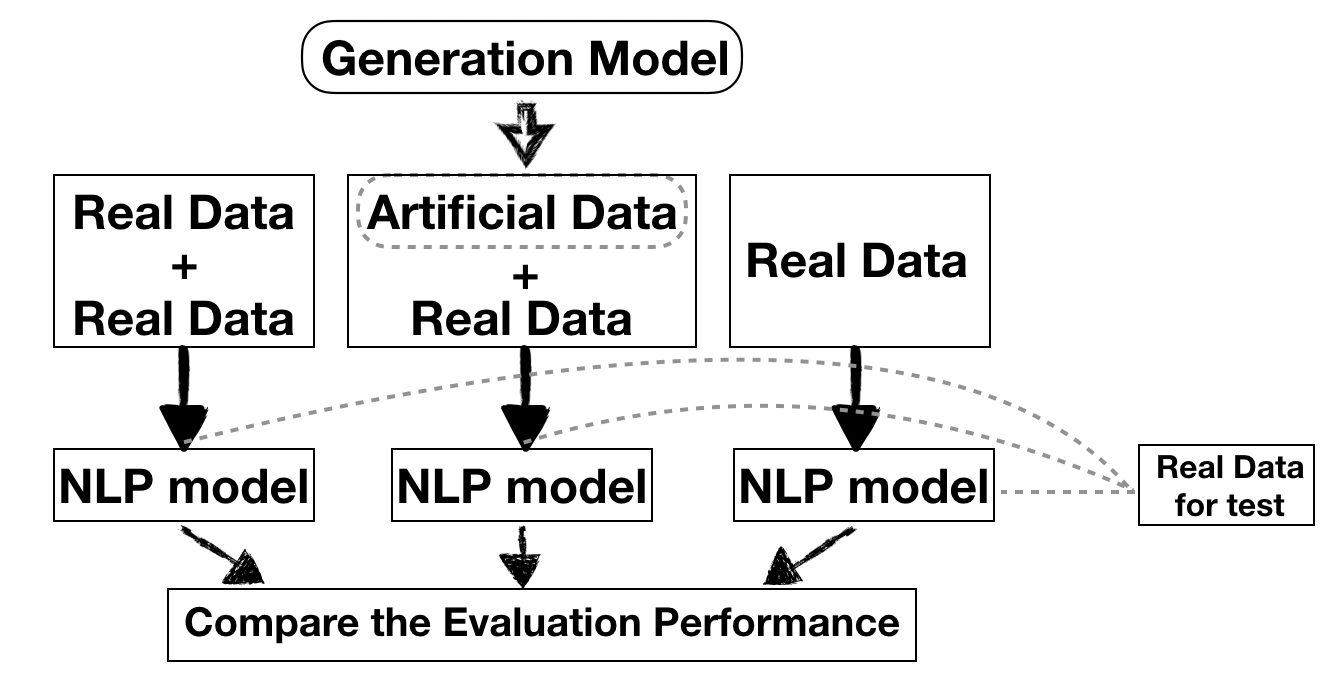}
  \caption{Our extrinsic evaluation procedure with real test data.}
  \label{fig:exeval}
\end{figure}

At the extrinsic evaluation step, we use \textit{generated data} as training data in a phenotype classification task and a temporal relation extraction task. For each task, we experiment with neural models. We compare performance of three models: one trained with real data, one trained using upsampled real data (the real dataset repeated twice) and one built using real data augmented with generated data for \textit{real test sets} (see Figure~\ref{fig:exeval}). Development sets are also real across setups. By upsampling the real data twice we create a baseline mimicking a very bad generation model simply reproducing the original data without adding any variation to it.

We further investigate the actual contribution of the artificial data to the classification process in experiments where we fully replace the real training data with the artificial training data for both neural and non-neural algorithms. Useful artificial data models should demonstrate similar performance results to real models. And, most importantly, those artificial data models should correctly preserve any differences between classification algorithms trained using the real data.

\section{Experimental Setup}\label{sec:exps}

In what follows, we describe the data used in experiments (Subsection~\ref{ssec:data}), details of generation models (Subsection~\ref{ssec:txt-gen-models}) and classification models (Subsection~\ref{ssec:txt-class-models}) we use. 

\subsection{Data}\label{ssec:data}

In our study we use EHRs from the publicly available MIMIC-III database \citep{johnson:mimic:2016,johnson:mimic-data:2016}. MIMIC-III contains de-identified clinical data of around 50K adult patients to the intensive care units (ICU) at the Beth Israel Deaconess Medical Center from 2001 to 2012. The dataset comprises several types of clinical notes, including discharge summaries, nursing notes, radiology and ECG reports. 

\par{\textbf{Text generation dataset.}} For the text generation experiments, we extract all the MIMIC-III discharge summaries of the patients with the 3 first diagnoses (ordered by their priority, represented by 2 first characters of each respective ICD-9 code) matching at least one sequence of the 3 first diagnoses for the patients from our phenotyping dataset (used later in our phenotype classification experiments). Thus, our text generation dataset do not contain the patients from the phenotyping dataset.

From all the extracted data we randomly select records of 126 patients for development purposes. This results in two subsets: \traingen and \devgen (see Table~\ref{table:stat-gen}). As our test sets we used parts of the phenotyping dataset (\testgen) and of the temporal relations dataset (\testgentemp) described below.

\begin{table}[!h]
\begin{center}
\scalebox{0.73}{
\begin{tabular}{c|cccc}
\toprule
set & \#, patient ID & \#, admission ID & \#, lines & \#, tok. \\ \midrule
\traingen & 9767 & 10926 & 1.2M & 20M\\
\devgen & 126 & 132 & 13K & 224K \\
\bottomrule
\end{tabular}}
\end{center}
\caption{\label{table:stat-gen} Statistics over \traingen, and \devgen. \# denotes number.}
\end{table}

Our preprocessing pipeline including sentence detection uses the spaCy-2.0.18 toolkit.\footnote{\url{https://spacy.io}}
We lowercase all texts. In addition, we replace dates with a placeholder \texttt{date}. We discard all the sentences with length under 5 words. 

\par{\textbf{Phenotyping dataset.}} In our text classification experiments we use the phenotyping dataset from  MIMIC-III database released by \citet{gehrmann:plos:2018}. Phenotyping is the task of determining whether a patient has a medical condition or is at risk for developing one. The dataset includes discharge summaries annotated with 13 phenotypes (e.g., advanced cancer, advanced heart disease, advanced lung disease, etc.)\footnote{\url{https://github.com/sebastianGehrmann/phenotyping}}

The phenotyping dataset used in our experiments contains 1,600 discharge summaries of 1,561 patients (around 180K sentences). We follow \citet{gehrmann:plos:2018} and randomly select 10\% and 20\% of this data for development and test purposes respectively (\devclass and \testclass). The rest 70\% is used as the test set for the generation experiments and as the training set for the phenotype classification experiments (\testgen).\footnote{Because of structural differences between MIMIC-III and MIMIC-II database that was initially used to collect the phenotyping dataset, we could not correctly identify text fields for records with duplicated admission IDs. We simply merged those records together giving preferences to annotations with a higher rate of  positive labels. This resulted in a small reduction of the initial dataset (less than 1\%).}

\par
\begin{figure}[h]
  \centering
  \includegraphics[width=0.95\hsize]{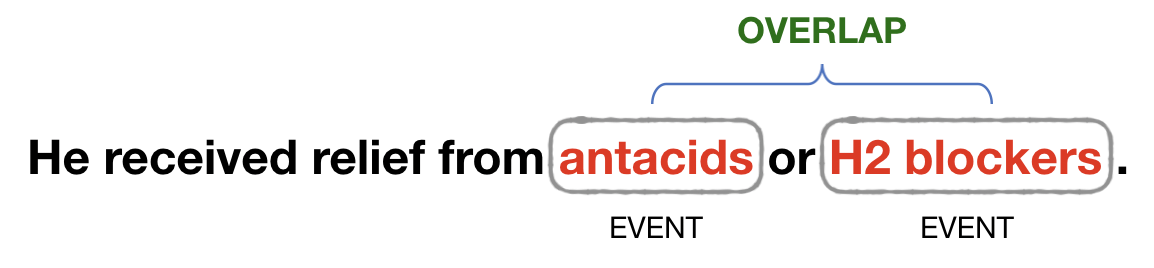}
  \caption{Example of an \overlap temporal relation (paraphrased).}
  \label{fig:overlap}
\end{figure}

{\textbf{Temporal relations dataset.}} In the temporal relation classification experiments, we use the data set from the 2012 i2b2 temporal relations shared task~\cite{sun2013evaluating}. The task focuses on determining the relative ordering of the events in medical history with respect to each other and to time expressions.
The dataset contains texts of discharge summaries from MIMIC-II. Various textual segments in these summaries are manually annotated for events (\texttt{EVENT}), time expressions (\texttt{TIMEX3}) and eight temporal relations between them (\texttt{TLINK}). In this study we focus only on detecting the presence of the most frequent \texttt{OVERLAP} temporal relation between events (33\% of the annotated relations). \overlap indicates that two related events happen almost the same time, but not exactly~\cite{Sun:2013aa} (see Figure~\ref{fig:overlap}).

The original training set includes 190 discharge summaries. We experiment with this dataset to demonstrate the transferability of our generation methodology. Hence, we do not modify our generation model but instead filter out the discharge summaries in the 2012 i2b2 dataset that overlap in their content with \traingen (according to the $\geq$ 10 sentences criteria). 

For the 2012 i2b2 data, we condition the generation using the textual segments annotated as \texttt{EVENT}. These could also be seen as binding elements of parts of longer text. Moreover, textual segments given in the input are mostly preserved in the generated output. The advantage of this approach is that in most of the cases we do not need to redo human annotation in the generated text because they are preserved if given in the input. Table~\ref{table:stat} reports the statistics of the original (\original) data versus the data (\reduced) for which the annotations are preserved.
 
10\% of the data is randomly selected for development purposes (\devtemp). The rest of the data is again used as the test data for the generation task and as the training data for the temporal classification models (\testgentemp). The test set provided with the 2012 i2b2 temporal relations shared task was used as is for temporal classification models (\testtemp).

\begin{table}[!h]
\begin{center}
\scalebox{0.73}{
\begin{tabular}{c|cccc}
\toprule
 & \#, docs & \#, lines & \#, tok. & \%, \overlap . \\ \midrule
\original & 190 & 7447 & 97K & 33.0\\
\reduced & 175 & 6762 & 89K & 33.6\\
\bottomrule
\end{tabular}}
\end{center}
\caption{\label{table:stat} Statistics over \testgentemp. The \original dataset corresponds to the one provided by the organisers. The \reduced dataset is the one for which the annotations are preserved by the generation model. \# denotes number.}
\end{table}

\subsection{Text Generation Models}\label{ssec:txt-gen-models}

In our text generation experiments we use the \transformer model, which generates text sentence by sentence. To ensure the semantic integrity of paragraphs resulting from the concatenation of generated sentences, we guide the generation with key phrases. Key phrases are extracted from each original paragraph of \traingen. For this, we use the \texttt{Rake} algorithm \citep{rake:2010}\footnote{The algorithm selects phrases composed of frequent content words co-occurring with other content words.} and take the highest scored 50\% per paragraph.
We further generate a paragraph sentence by sentence using as inputs only those extracted key phrases that are present in each particular sentence. This results in approximately 2.4 key phrases with an average length of 1.7 words per sentence (as computed for \traingen).\footnote{We used the implementation available at \url{https://github.com/csurfer/rake-nltk}} Boundaries of key phrases in the input to models are fixed by a reserved token. During training, the model is learned to restore real text from key phrases, basically by filling the gaps between those key phrases.

We trained our \transformer models as provided by the OpenNMT toolkit \citep{opennmt:2017} with default parameters. In \traingen we replaced all the words with frequency 1 with a placeholder. This resulted in a vocabulary of around 50K words. Each model was trained for 30K epochs.\footnote{We noticed that this quantity of epochs is necessary for stabilization of the model perplexity.} Outputs are produced with the standard beam decoding procedure with a beam size of 5.

\begin{table*}[t!]
\begin{center}
\scalebox{0.7}{
\begin{tabular}{c|c|p{17.3cm}}
\toprule
\multicolumn{3}{c}{\testgen}\\
\midrule
1 & \gen & a ct was obtained which revealed a very poor study but \underline{no evidence of} a brain injury .\\
& \real & ct was a poor study but \underline{did not reveal} a brain injury .\\
\midrule
2 & \gen & he had \underline{a walk of losing blood} .\\
& \real & she is unable \underline{to walk without losing blood} .\\
\midrule
\multicolumn{3}{c}{\testgentemp}\\
\midrule
3 & \gen & he was treated with increasing doses of \underline{rosuvastatin and atorvastatin} .\\
& \real & he has been on increasing doses of \underline{rosuvastatin receiving atorvastatin in addition} on a basis .\\
\midrule
4 & \gen & he was started on ibuprofen and \underline{his wife back pain} was improved .\\
& \real & the patient was initially treated with ibuprofen which was stopped after \underline{his back pain} improved .\\
\bottomrule
\end{tabular}}
\end{center}
\caption{\label{generation-examples} Examples of real and generated text. The underlined text highlights ``good'' (examples 1 and 3) or ``bad'' (examples 2 and 4) modifications. All sentences have been paraphrased.
}
\end{table*}

\subsection{Models for Phenotype Classification}\label{ssec:txt-class-models} 

For the phenotype classification task, we train two standard NLP models:
\begin{enumerate}
\item \textbf{Convolutional Neural Network (CNNs) model} inspired by \cite{kim:corr:2016}. The CNN model is built with 3 convolutional layers with window sizes of 3, 4 and 8 respectively. The word embedding dimensionality is 300, both convolution layers have 100 filters. The size of the hidden units of the dense layer is 100. We also use a dropout layer with a probability of 0.5. The network is implemented using the Pytorch toolkit\footnote{\url{https://pytorch.org}} with the pre-trained GloVe word embeddings \cite{pennington2014glove}.

\item \textbf{Word-level bag-of-words (BoW) model} trained with the Naive Bayes (NB) algorithm. We applied the MultinomialNB implementation from \texttt{Scikit-learn}~\cite{scikit-learn}.
\end{enumerate}

We cast the task as a binary classification task and evaluate the detection of each phenotype computing the F1-score of the positive class.
 
\subsection{Models for Temporal Relations Extraction}
 
Inspired by the SOTA approaches for the task~\cite{tourille-EtAl:2017:Short}, we build a Bidirectional Long Short-Term Memory (BiLSTM) classifier~\cite{hochreiter}. The BiLSTM model is constructed with two hidden layers of opposite directions. The size of hidden LSTM units is 512. We use a dropout layer before the output layer with a probability of 0.2 and the concatenation of the last hidden states of both layers goes into the ouput layer. We train our network with the Adam~\cite{kingma2014adam} optimization algorithm with a batch size of 64 and a learning rate of 0.001. We use again the pre-trained GloVe word embeddings. The classifier is implemented using Pytorch. As for a non-neural model, we use again the NB model as for the phenotype classification task.

We cast the task as a binary classification task (for each event-event pair, classify as \overlap or not) and evaluate the result by computing the F1-score of the positive decision.

\begin{table}[t!]
\begin{center}
\scalebox{0.7}{
\begin{tabular}{c||c|c|p{2.4cm}}
\toprule
& \rouge & \bleu & avg. sent. $l$ (\gen./\real)\\\midrule
\testgen & 67.74 & 40.62 & 13.27 / 17.50 \\
\testgentemp & 48.47 & 20.91 & 18.61 / 16.81 \\
\bottomrule
\end{tabular}}
\end{center}
\caption{\label{table:intr-eval} Qualitative evaluation and average sentence lengths. 
}
\end{table}

\section{Experimental Results}\label{sec:exp-res}

In this section we present results of our experiments, first of the intrinsic evaluation of the quality of generated text (Section \ref{ssec:intr-eval}) and then of the extrinsic evaluation of its utility for NLP (text classification and temporal relation extraction tasks, Section \ref{ssec:extr-eval}).

\begin{table*}[t!]
\begin{center}
\scalebox{0.60}{
\begin{tabular}{c||c|c|c|c|c|c|c|c|c|c|c|c|c||c}
\toprule
& \rotatebox[origin=c]{90}{Obesity}& \rotatebox[origin=c]{90}{Non Adherence}&
\rotatebox[origin=c]{90}{Developmental.Delay.Retardation} &\rotatebox[origin=c]{90}{Adv. Heart Disease} & \rotatebox[origin=c]{90}{Adv. Lung Disease}& \rotatebox[origin=c]{90}{Schizo and other Psych. Disorders}& \rotatebox[origin=c]{90}{Alcohol Abuse}& \rotatebox[origin=c]{90}{Other Substance Abuse}& \rotatebox[origin=c]{90}{Chr. Pain Fibromyalgia}& \rotatebox[origin=c]{90}{Chr. Neurological Dystrophies} & \rotatebox[origin=c]{90}{Adv. Cancer}& \rotatebox[origin=c]{90}{Depression} & \rotatebox[origin=c]{90}{Dementia}& \rotatebox[origin=c]{90}{\textbf{avg.}}  \\
\midrule
freq, \% & 8 & 9 & 3 &17 &10 & 18 & 12 & 10 & 20 & 23 & 10 & 29 & 7 & \\
\midrule
& \multicolumn{14}{c}{CNN}\\
\midrule
\real + \gen  & 0.3257 & 0.3394&\bf0.3636 & \bf0.6384 & 0.5333 &\bf0.3664 & 0.7428 & \bf 0.5714 & 0.3846 & \bf 0.5574 & \bf 0.6173 & 0.4373 & 0.5714 & \bf 0.4961  \\
\real & \bf 0.3789 & \bf 0.3589  & 0.2500 & 0.6019 & 0.5085 & 0.2909 & 0.7200 & 0.4912  & 0.4000 & 0.4782 & 0.5567 & \bf 0.4623 & 0.5667 &  0.4665 \\
2 $\times$ \real & 0.3636 & 0.3333  & 0.2857 & 0.5347 & \bf 0.5758 & 0.3057 & \bf 0.7435 & 0.4789 & \bf 0.4040 & 0.4580  & 0.5667 & 0.4162 & \bf 0.6341 & 0.4692 \\
\midrule
& \multicolumn{14}{c}{}\\
\midrule
\gen & 0.2500 & 0.3656  & 0.2000 & 0.4667 & 0.5574 & 0.3221 & 0.7297 & 0.4478 & 0.3978 & 0.4564 & 0.6575 & 0.4598 & 0.3273 & 0.4337  \\
\nometa & 0.1365 & 0.2443  & 0.0252 & 0.5200 & 0.1429 & 0.2978 & 0.2581 & 0.1914 & 0.3781 & 0.3740 & 0.3778 & 0.4262 & 0.0800 & 0.2656   \\
\midrule
& \multicolumn{14}{c}{NB}\\
\midrule
\real& 0.2000 & 0.4722  & 0.0000 & 0.5812 & 0.4838 & 0.5614 & 0.6756 & 0.5000  & 0.4109 & 0.5270 & 0.6779 & 0.5700 & 0.3846 & 0.4650   \\
\hline
\gen & 0.2424 & 0.4719  & 0.0000 & 0.5893 & 0.4687 & 0.5000 & 0.6506 & 0.4594 & 0.4022 & 0.5122 & 0.6562 & 0.5391 & 0.3125 & 0.4465   \\
\nometa & 0.1407 & 0.1984 & 0.0447 & 0.3022 & 0.2108 & 0.2857 & 0.2367 & 0.1723 & 0.3284 & 0.3815 & 0.2032 & 0.4398 & 0.1039 & 0.2345   \\
\bottomrule
\end{tabular}}
\end{center}
\caption{\label{table:text-class-res} Phenotyping results for CNN and Naive Bayes (NB), \testclass. Best performing models for CNN data augmentation experiments are highlighted in bold. We report results for the models trained with: \real data augmented with generated \gen data, \real data only, 2 $\times$ \real data upsampled twice, \gen data only, \nometa data without traces of the input real data.
}
\end{table*}

\subsection{Intrinsic Evaluation}\label{ssec:intr-eval} 

Table~\ref{table:intr-eval} shows the intrinsic evaluation results for both generated \testgen and \testgentemp. The \bleu and \rouge are computed between the original text (the one used to extract key phrases) and the generated text. We also compare the average lengths sentences for those two texts.

As expected, automatic evaluation scores show that for both test sets our model generates context preserving pieces of the real text from the input (e.g., $\rouge=67.74$ for \testgen, $\rouge=48.47$ for \testgentemp). The proximity of average lengths of sentences for the generated text and the real text supports this statement.

As automatic metrics perform only a shallow comparison, we also manually reviewed a sample of texts. In general, most of the generated text preserves the main meaning of the original text adding or dropping some details. Incomprehensible generated sentences are rare. 

Table~\ref{generation-examples} shows examples of the generated text for both datasets. In examples 1 and 3, \transformer generates text with a meaning very close to the original one (e.g., \textit{no evidence of}~$\approx$ \textit{did not reveal}, for \testgen). Examples 2 and 4 are ``bad'' modifications. In general, such examples are infrequent. For instance in Example 2, the real phrase \textit{unable to walk without losing blood} is incorrectly modified into \textit{a walk of losing blood}. However, the main sense of losing blood is preserved.

Overall, our observations indicate that the generation methodology successfully adapts to changes in generation conditions.

\subsection{Extrinsic Evaluation}\label{ssec:extr-eval} 

\par{\textbf{Phenotype Classification.}} Table~\ref{table:text-class-res} shows results of our text classification experiments. They indicate that the artificial training data used as complimentary to the real training data is in general beneficial for the CNN model (e.g., av. F-score=$0.50$ for \real + \gen $> 0.47$ for \real). \real+\gen setup also outperforms the model trained using larger volume data, where the training data was repeated two times (2 $\times$ \real). Overall, \real+\gen outperforms \real for 9 phenotypes out of 13 with an average $\Delta$F-score=$0.06$, while 2 $\times$ \real for 6 phenotypes with an average $\Delta$F-score=$0.04$ only.

\begin{table*}[t!]
\begin{center}
\scalebox{0.68}{
\begin{tabular}{c||c|p{17.3cm}}
\toprule
 & F-score & \textbf{Features (words)} \\
\midrule
\real & 0.5614 & \textbf{chest}, 20, given, 11, hours, time, history, admission, continued, capsule, needed, 25, disease, refills, follow, negative, started, status, disp, days, release, discharge, ml, stable, hct, prior, dr, showed, 40, fax, neg, telephone, likely, 15, glucose, wbc, home, renal, care, seen, iv, 24, acute, urine, post, noted, artery, 14, year, unit, tube, inr, bid, 50, \textbf{edema}, units, plt, insulin, known, course, \textbf{pulmonary}, mild, did, dose\\
\midrule
\gen & 0.5000 & follow, 12, fax, renal, admission, care, telephone, prior, artery, bid, acute, dr, unit, known, time, post, likely, seen, neg, discharge, iv, insulin, tube, units, admitted, placed, year, 11, 25, 13, \textbf{pulmonary}, urine, dose, delayed, mild, chronic, transferred, \textbf{edema}, lower, pressure, heart, course, fluid, failure, ventricular, aortic, abdominal, 50, discharged, medications, valve, evidence, noted, increased\\
\midrule
\nometa & 0.2857 & blood, day, mg, 10, 07, date, pt, 10pm, refills, 100, 20, tablet, needed, started, \textbf{ct}, plt, 12, 30, inr, 11, 25, 13, dr, times, 50, sig, 213, 24, patient, daily, 40, 500, telephone, release, transferred, negative, discharged, 81, follow, final, admitted, 15, 30pm, time, fax, hours, delayed, normal, placed, history, 20am, seen, \textbf{breath}, 00, did, 18, 15pm, evidence, 80, admission, consulted, home, wbc, po, hct, bedtime, \textbf{shortness} \\
\bottomrule
\end{tabular}}
\end{center}
\caption{\label{feature-nb-tran} Top-30 words contributing the most to the \texttt{Advanced Lung Disease} phenotype detection using Naive Bayes.
}
\end{table*}

\begin{figure}[h]
  \centering
  \includegraphics[width=0.95\hsize]{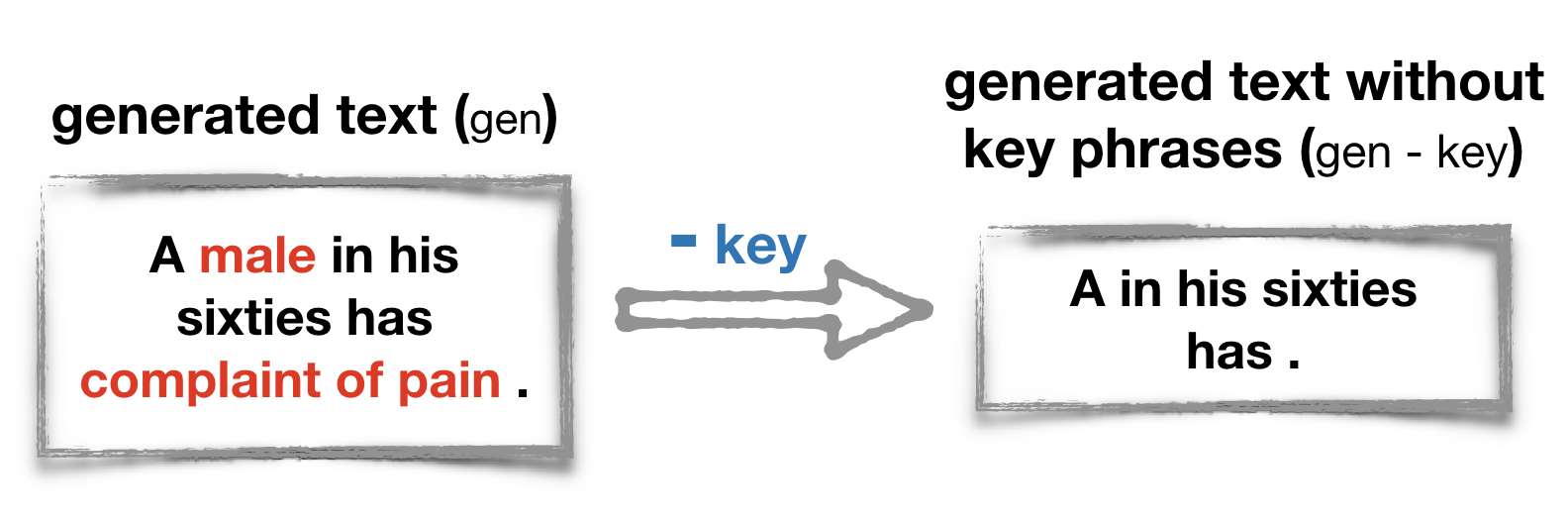}
  \caption{Example of creating \gen\nometa data -- the generated text without traces of input data (paraphrased)}
  \label{fig:gen-key}
\end{figure}

To get further insights into the actual informativeness of the generated data, we study the performance of both CNN and NB in a series of setups where the artificial training data fully replace the real training data. To be more precise, we study: (a) \gen setup, where the full generated data with traces of input key phrases are used as the training data; and (b) \nometa setup, where the generated text without traces of input data is used as the training data~(see Figure~\ref{fig:gen-key}). The results of these experiments are in Table~\ref{table:text-class-res}, lower part. They show that average performances of \gen and \real tend to be comparable for each algorithm (e.g., $\Delta$ avg. F-score=0.03 for both CNN and NB). The \nometa setup results in a significant performance drop (of F-score=0.2 on average). However, the \nometa text still potentially bears some relevant information that allows both CNN and NB have comparable performance for this setup.

Taking advantage of the easy interpretability of the NB model, we analyse the words that contribute the most to classification decisions (highest likelihoods given the positive class) for the \texttt{Adv. Lung Disease} as a an example of a phenotype with an average frequency for the dataset. Table~\ref{feature-nb-tran} displays those words in order of importance for \real, \gen and \nometa. As expected, for \real and \gen with higher F-score values, there are more relevant medical terms: e.g., \textit{pulmonary} and \textit{chest}. For \nometa, there are words more distantly related to the phenotype: e.g., \textit{ct} and \textit{breath}.

\begin{figure}[h]
  \centering
  \includegraphics[width=0.95\hsize]{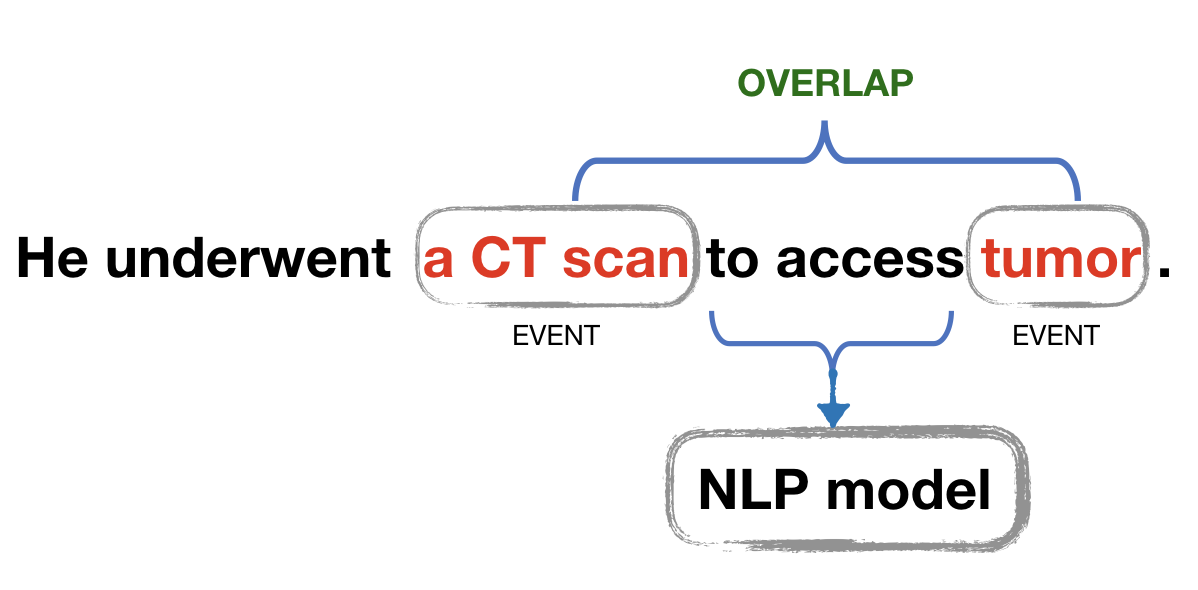}
  \caption{Example of an input to our models for temporal relations extraction -- a text span that links the two events (paraphrased).}
  \label{fig:tempeval}
\end{figure}

\par{\textbf{Temporal Evaluation.}} For the i2b2 dataset, we focus only on the evaluation for the \overlap temporal relation between events as the most well-represented group. Inspired by the SOTA solutions for the temporal relations extraction task~\cite{tourille-EtAl:2017:Short}, we provide only the text spans that link the two events as inputs to our models. This setup is particularly beneficial to assess the utility of the generated text (see Figure~\ref{fig:tempeval}). As mentioned earlier, for this dataset we guide the text generation with event text spans. Thus, for this setup, we take only the text between those real text spans essentially copied from the input. This allows us to better assess the utility only of what was generated.\footnote{However, it should be noted here that the generated text between two events may still contain other event spans copied from the input, especially for the cases when events are in different sentences.}

Table~\ref{tmp-trans} reports results for our experiments with the i2b2 dataset. They are similar to the ones performed for the phenotyping dataset. Note that we reduce the initial training set provided by the task due to particularities of our generation procedure. In our data augmentation experiments we add this reduced generated data to all the provided real training data (\real \original).

\begin{table}[!htbp]
\begin{center}
\scalebox{0.73}{
\begin{tabular}{c||c}
\toprule
& BiLSTM \\
\hline
\real \original + \gen & \bf 0.6217\\
\real \original & 0.5896\\
2 $\times$ \real \original & 0.5803 \\
\midrule
& \\
\midrule
\gen & 0.5138 \\
\real \reduced & 0.5312 \\
\midrule
& NB \\
\midrule
\gen &  0.5769 \\
\real \reduced & 0.5024\\
\bottomrule
 \end{tabular}}
\end{center}
\caption{\label{tmp-trans} Temporal relations extraction for \overlap for CNN and Naive Bayes (NB), \testtemp. Only the real/generated text between events serves as input. Best performing models for data augmentation experiments are highlighted in bold. We report results for the models trained using: \real \original training data from the i2b2 task augmented with the generated \gen data, \real \original data only, 2 $\times$ \real \original data upsampled twice, \real \reduced data only, \gen data only. 
}
\end{table}

The results show that \real \original + \gen (F-score=$0.62$) outperforms the \real setup (F-score=$0.59$), as well as the upsampled setup (2 $\times$ \real \original, F-score=$0.58$). This confirms the utility of our data augmentation procedure for the BiLSTM model. Results for \gen and \real \reduced are again comparable for BiLSTM. For NB, we even observe an improvement of $\Delta$F-score=$0.08$ for \gen as compared to \real \reduced for NB. This may be explained by a stronger semantic signal in the generated data. Overall, our results demonstrate the potential of developing a model that would generate artificial medical data for a series of NLP tasks.

\section{Discussion}\label{sec:discussion} 

Our study is designed as a proof-of-concept and the main objective of this work is to study the utility of using SOTA approaches for generating artificial EHR data and to evaluate the impact of using this to augment real data for common NLP tasks in the clinical domain. Our results are promising. From a preliminary manual analysis, most meaning is preserved in the generated texts. For both extrinsic evaluation tasks (phenotype classification, and temporal relation classification), using generated text to augment real data in the training phase improved results. Moreover, for both tasks, results using only generated data was comparable to those using only real data, further indicating usefulness.

To our knowledge, this is the first study looking at the problem of generating longer clinical text, and that is extrinsically evaluated on two downstream NLP tasks. Although the MIMIC data is comprehensive, it represents a particular type of clinical documentation from an ICU setting, in further work we plan to extend to other clinical domains.

If artificial data was to be used for further downstream tasks, particularly those that are intended to support secondary uses in a clinical research setting, further analysis is needed to assess the clinical validity of the generated text. This would require domain expertise. For instance, the temporal relation classification problem imposes different constraints as compared with the document classification task, which might require other approaches for designing the text generation models. Moreover, other temporal information representation models have been proposed in other studies, for other use-cases, such as the \texttt{CONTAINS} relation in the THYME corpus \cite{styler2014}. In future studies, we will invite clinicians to review the generated text with a focus on clinical validity aspects, as well as study further downstream NLP tasks. We will also study additional alternative metrics for intrinsic evaluation, such as the modified CIDEr metric proposed by \citet{lee2018nlg}.

\section{Conclusion}\label{sec:concl}
In this work, we attempt to generate artificial training data for two downstream clinical NLP tasks: text classification and temporal relation extraction. 
We propose a generic methodology to guide the generation in both cases. Our experiments show the utility of artificial data for neural NLP models in data augmentation setups. Our generation methodology holds promise for the development of a more universal approach that will allow medical text generation for an even wider range of biomedical NLP tasks. We also plan to further investigate the validity and utility of artificial data. We think thus, that artificial data generation is an approach that has the potential to solve current data accessibility issues associated with biomedical NLP.

\section*{Acknowledgments}
This work was partly funded by EPSRC Healtex Feasibility Funding 
(Towards Shareable Data in Clinical Natural Language Processing: 
Generating Synthetic Electronic Health Records). The third author has received support from the Swedish Research Council (2015-00359), and the Marie 
Sk\l{}odowska Curie Actions, Cofund, Project INCA 600398. We would like to thank the anonymous reviewers for their helpful comments. 

\bibliography{acl2019}
\bibliographystyle{acl_natbib}

\end{document}